\documentclass[letterpaper]{article} 
\usepackage{aaai23}  
\usepackage{times}  
\usepackage{helvet}  
\usepackage{courier}  
\usepackage[hyphens]{url}  
\usepackage{graphicx} 
\usepackage{natbib}  
\usepackage{caption} 
\usepackage{algorithm}
\usepackage{algorithmic}
\usepackage{amsmath}
\usepackage[table,xcdraw]{xcolor}
\usepackage[normalem]{ulem}
\usepackage{subfigure}
\useunder{\uline}{\ul}{}
\usepackage{newfloat}
\usepackage{listings}
\DeclareCaptionStyle{ruled}{labelfont=normalfont,labelsep=colon,strut=off} 
\floatstyle{ruled}
\newfloat{listing}{tb}{lst}{}
\floatname{listing}{Listing}
\title{Auto-weighted Multi-view Clustering for Large-scale Data}
\author{Xinhang Wan,\textsuperscript{\rm 1} Xinwang Liu,\textsuperscript{\rm 1}\footnote{Corresponding author} Jiyuan Liu,\textsuperscript{\rm 1} Siwei Wang ,\textsuperscript{\rm 1} Yi Wen,\textbf{\textsuperscript{\rm 1}} \\ Weixuan Liang,\textsuperscript{\rm 1} En Zhu,\textsuperscript{\rm 1} Zhe Liu,\textsuperscript{\rm 2} Lu Zhou\textsuperscript{\rm 2}}
\affiliations{
    \textsuperscript{\rm 1}College of Computer, National University of Defense Technology, Changsha, China\\
    \textsuperscript{\rm 2}College of Computer Science and Technology, Nanjing University of Aeronautics and Astronautics, China
    
    \{wanxinhang, xinwangliu, liujiyuan13, wangsiwei13, wenyi21, weixuanliang, enzhu\}@nudt.edu.cn, \{zhe.liu, lu.zhou\}@nuaa.edu.cn
}

\usepackage{bibentry}

\begin{document}

\maketitle

\begin{abstract}
Multi-view clustering has gained broad attention owing to its capacity to exploit complementary information across multiple data views. Although existing methods demonstrate delightful clustering performance, most of them are of high time complexity and cannot handle large-scale data. Matrix factorization-based models are a representative of solving this problem. However, they assume that the views share a dimension-fixed consensus coefficient matrix and view-specific base matrices, limiting their representability. Moreover, a series of large-scale algorithms that bear one or more hyperparameters are impractical in real-world applications. To address the two issues, we propose an auto-weighted multi-view clustering (AWMVC) algorithm. Specifically, AWMVC first learns coefficient matrices from corresponding base matrices of different dimensions, then fuses them to obtain an optimal consensus matrix. By mapping original features into distinctive low-dimensional spaces, we can attain more comprehensive knowledge, thus obtaining better clustering results. Moreover, we design a six-step alternative optimization algorithm proven to be convergent theoretically. Also, AWMVC shows excellent performance on various benchmark datasets compared with existing ones. The code of AWMVC is publicly available at https://github.com/wanxinhang/AAAI-2023-AWMVC.
\end{abstract}
\section{Introduction}
With the rapid development of multimedia techniques, data can be described from various modalities \cite{XiaWGZG22, KGESymCL}. For instance, an object could be recognized by an intelligent robot from its eyes (image), ears (transform sound into texts), and past knowledge (knowledge graph). How to uncover items' intrinsic structure and label them is crucial in many applications \cite{zhang2022efficient,yang2022interpolation, XiaWGYG21}, such as the recommender system and decision support system. Multi-view clustering, which explores complementary information among views and discovers the underlying structure for clustering, is an effective method to conduct a clustering process for data from multiple sources. 

To the best of our knowledge, existing multi-view clustering algorithms can be divided into four categories including multi-view subspace clustering \cite{7410839,9399655,liu2022efficient}, multiple kernel clustering \cite{9653838,LiangTNNLS, ZJPACMMM, 10.1145/3503161.3547864}, graph-based methods \cite{2020MultiLiang,tang2020,liu-1,liu-3,XGWGCT2022}, and matrix factorization-based methods \cite{9170204,Gao2019MultiviewLM}. Under the assumption that a linear combination of data samples can reconstruct themselves, multi-view subspace clustering obtains a reconstruction matrix upon a self-expressive framework. For instance, \cite{8099944} simultaneously explores the latent representation of data samples and the underlying complementary information to attain a more comprehensive reconstruction matrix. Multiple kernel clustering \cite{9556554} always seeks a common cluster assignment matrix by jointly maximizing the partition matrix and kernel coefficients. As pointed out by \cite{9146384}, graph-based methods rely on spectral clustering, which performs eigendecomposition upon the Laplacian matrix to partition the data cloud with a linear/nonlinear relationship.

Despite the above three categories of algorithms displaying acceptable clustering performance, they encounter the square or cubic time complexity, which limits the applicability in large-scale data. To handle the issue, a series of researches \cite{liu2013multi, Gao2019MultiviewLM} based on non-negative matrix factorization (NMF) appear due to its low time complexity. NMF-based multi-view clustering factorizes the original features into two components, i.e., a consensus coefficient matrix and view-specific base matrices. For example, \cite{8030316} proposes a method to reduce the redundancy using a diversity term and solves the resultant optimization problem in linear execution time. However, the algorithm puts up with two hyperparameters, restraining the practicability in many applications. Based on the observation that the non-negative constraint leads to a less discriminatory embedding, \cite{9710821} removes the restriction and proposes an MF-based algorithm. However, it assumes that the latent embedding of multiple views is under a fixed latent dimension, which harms the expressiveness of the model and the ability to extract complementary information between views. 

 We propose auto-weighted multi-view clustering for large-scale data (AWMVC) to address the abovementioned drawbacks. To enhance the data embedding and gain complementary information among views, we obtain various coefficient matrices learned by base matrices under diverse dimensions, then integrate the coefficient matrices into a consensus one. It is worth noting that the diverse coefficient matrices can guide a more robust consensus item. As feedback, the consensus one promotes the quality of coefficient matrices likewise. In addition, the proposed method is parameter-free. Considering that the quality of each specific dimension exists differences, we automatically measure the importance of each one by the corresponding contributions to the final result. We develop a six-step alternate optimization algorithm with proven convergence to solve the resulting optimization problem. The linear time complexity and parameter-free property enable AWMVC to handle large-scale data. Compared with state-of-the-art multi-view clustering algorithms on various benchmark datasets, AWMVC displays effectiveness and superiority. In general, the contributions of the paper are summarized as follows:
 \begin{enumerate}
\item We remove the non-negative constraint of NMF and obtain coefficient matrices with view-specific base matrices of different dimensions, then integrate the coefficient matrices into a consensus one and combine these two separate steps into a unified one, which filters redundant information and attains a more comprehensive knowledge.
\item The proposed algorithm can be carried out in linear time complexity with no hyperparameters, which is suitable for large-scale applications. The importance of each part is automatically tuned by its corresponding contribution to the final result. Therefore, the clustering result is superior with more discriminatory information. 
\item We develop a six-step alternative optimization algorithm with proven convergence. To validate the effectiveness of AWMVC, we conduct extensive experiments on various datasets. The results demonstrate the efficiency and excellent performance of the proposed method. 
\end{enumerate}

\section{Related Work}
In this section, we briefly overview MF-based single-view clustering and MF-based multi-view clustering separately.

\subsection{MF-based Single-view Clustering}
Given a single-view dataset $\mathbf{X}=\left[\mathbf{x}_{1}, \ldots, \mathbf{x}_{n}\right] \in \mathrm{R}^{d \times n}$, where $d$ and $n$ denote the number of features and samples respectively, NMF seeks to factorize data matrix into two non-negative components, i.e., base matrix $\mathbf{H}\in \mathrm{R}^{d \times k}$ and coefficient matrix $\mathbf{Z}\in \mathrm{R}^{k \times n}$ with $k$ indicating the cluster numbers. The goal of NMF is to approximate the product of $\mathbf{H}$ and $\mathbf{Z}$ to $\mathbf{X}$ as follows

 \begin{equation}\label{ORIGINAL_NMFSVC}
\begin{aligned}
\min _{\mathbf{H} \geq \mathbf{0}, \mathbf{Z} \geq \mathbf{0}} f(\mathbf{X}, \mathbf{H Z}),
\end{aligned}
\end{equation}
where $f(\cdot)$ is a loss function. Most papers measure the loss between the two items by adopting the Frobenius norm \cite{liu2013multi,9163943}. Based on the observation that the orthogonality constraint on NMF leads to rigorous clustering interpretation \cite{2006Orthogonal}, a large number of researchers have investigated the NMF of orthogonality constraint. By imposing orthogonality constraints on base matrices, Eq. \eqref{ORIGINAL_NMFSVC} can be formulated into

\begin{equation}\label{ORTHO_NMFSVC}
\begin{aligned}
\min _{\mathbf{H} \geq \mathbf{0}, \mathbf{Z} \geq \mathbf{0}}\|\mathbf{X}-\mathbf{H Z}\|_{F}^{2} \quad \text { s.t. } \mathbf{H}^{\top} \mathbf{H}=\mathbf{I}_{k}.
\end{aligned}
\end{equation}

Noticing that the non-negative constraint produces less discriminatory information, \cite{9710821} removes the restriction and proposes an MF-based method, as shown
\begin{equation}\label{ORTHO_MFSVC}
\begin{aligned}
\min _{\mathbf{H}, \mathbf{Z}}\|\mathbf{X}-\mathbf{H Z}\|_{F}^{2} \quad \text { s.t. } \mathbf{H}^{\top} \mathbf{H}=\mathbf{I}_{k}.
\end{aligned}
\end{equation}

After getting the coefficient matrix $\mathbf{Z}$, a traditional clustering method like $k$-means is conducted to obtain the final cluster assignments.

\subsection{MF-based Multi-view Clustering}
Given the multi-view data $\left\{\mathbf{X}^{(v)}\right\}_{v=1}^{V}$, in which $\mathbf{X}^{(v)} \in \mathbf{R}^{d_{v} \times n}$ and $d_{v}$ denotes its feature dimension, MF-based multi-view clustering aims to seek a consensus matrix $\mathbf{Z}$ to extract information among views. One of the most representative formulas is as follows \cite{8030316}
\begin{equation}\label{ORI_MFMVC}
\begin{aligned}
&\min _{\mathbf{H}, \mathbf{Z}}\sum_{v=1}^{V}\frac{1}{2}\left\|\mathbf{X}^{(v)}-\mathbf{H}^{(v)} \mathbf{Z}^{(v)}\right\|_{F}^{2}+\lambda \sum_{v \neq w} \operatorname{DIVE}\left(\mathbf{Z}^{(v)}, \mathbf{Z}^{(w)}\right) \\
&\text { s.t. }  1 \leq v, w \leq V, \mathbf{H}^{(v)}\geq \mathbf{0}, \mathbf{Z}^{(v)}\geq \mathbf{0}, \mathbf{Z}^{(w)} \geq \mathbf{0},
\end{aligned}
\end{equation}
where $\operatorname{DIVE}\left(\cdot,\cdot\right)$ is to ensure the diversity of the representation matrices and $\lambda$ is the balanced hyperparameter. The consensus matrix $\mathbf{Z}$ is acquired via the average value of $\mathbf{Z}^{(v)}$. On the contrary, some methods have been proposed to directly attain a common underlying matrix across views, shown as follows

\begin{equation}\label{ORTHO_MFMVC}
\begin{aligned}
&\min _{\mathbf{H}, \mathbf{Z}} \sum_{v=1}^{V}\frac{1}{2}\left\|\mathbf{X}^{(v)}-\mathbf{H}^{(v)} \mathbf{Z}\right\|_{F}^{2}+\lambda\Phi(\mathbf{H}, \mathbf{Z}) \\
&\text { s.t. } {\mathbf{H}^{(v)}}^{\top} \mathbf{H}^{(v)}=\mathbf{I}_{k},
\end{aligned}
\end{equation}
where $\Phi(\cdot)$ are some regularization terms on $\mathbf{H}$ and $\mathbf{Z}$. 

Although these above methods display acceptable performance, they suffer from two drawbacks that are hard to solve. First, most MFMVC algorithms decompose data matrices from multiple views into a fixed-dimension embedding, which ruins the representational power of the model. Meanwhile, insufficient data acquisition cannot ensure complementary knowledge among views. As a consequence, the final partition matrix fails to extract adequate information. Second, to balance the weights between the loss function and regularization terms, hyperparameters are inevitable in many methods, limiting their extensions to large-scale applications. In the next section, we propose a novel auto-weighted multi-view clustering method for large-scale data to handle these problems.

\section{Methodology}
In this section, we first provide the objective formulation of AWMVC, then offer a six-step alternate optimization algorithm to solve the resultant problem. After that, we will discuss its time complexity and convergence.

\subsection{The Proposed Method}
As mentioned above, current MFMVC methods are presented as Eq. \eqref{ORTHO_MFMVC}. Although they demonstrate advantages among exiting MVC algorithms, there are still some drawbacks they fail to solve simultaneously. We observe that MFMVC encounters two main problems. First, the current MFMVC maps original features of various views into a fixed dimension. However, data matrices from multiple sources usually belong to specific latent dimensions, and the clustering result significantly depends on the dimensions. Most researchers ignore this shortcoming and merely set it equal to the number of clusters to avoid selecting the dimensions from a range of values. In contrast, others tackle this problem by introducing a hyperparameter. To deal with this issue, we project the original feature matrix of each view into several dimensions and automatically balance their weights according to their corresponding contributions. Second, state-of-the-art methods encounter one or more hyperparameters, preventing them from being applied to large-scale applications. Consequently, we eliminate the regularization term, and the optimization goal is as follows 
\begin{equation}\label{diverse_alpha_MFMVC}
\begin{aligned}
&\min _{\boldsymbol{\alpha},\mathbf{H}, \mathbf{Z}} \sum_{p=1}^{m}\sum_{v=1}^{V} \frac{1}{2}\alpha_p^2 \left\|\mathbf{X}^{(v)}-\mathbf{H}_{p}^{(v)} \mathbf{Z}_{p}\right\|_{F}^{2} \\
&\text { s.t. } \boldsymbol{\alpha}^{\top} \mathbf{1}=1, \boldsymbol{\alpha}\geq\mathbf{0}, {\mathbf{H}_{p}^{(v)}}^{\top} \mathbf{H}_{p}^{(v)}=\mathbf{I}_{k},
\end{aligned}
\end{equation}
where $m$ denotes the total number of latent embedding of each view, and $\alpha_p$ indicates the weight of $\mathbf{Z}_{p}$.

By solving Eq. \eqref{diverse_alpha_MFMVC}, we obtain $\mathbf{Z}_{p}$ with different dimensions. How to fuse them efficiently is a new issue. We address this problem via mapping them into a $k$-dimension space by a rotation matrix. The separate process between coefficient matrices learning and the consensus one learning may result in a sub-optimal result, so we unify them into a unified step as
\begin{equation}\label{AWMVC}
\begin{aligned}
&\min  \sum_{p=1}^{m} \sum_{v=1}^{V} \frac{1}{2}\alpha_p^2 \left\|\mathbf{X}^{(v)}-\mathbf{H}_{p}^{(v)} \mathbf{Z}_{p}\right\|_{F}^{2} \\
&-\sum_{p=1}^{m} \beta_p\operatorname{Tr}\left({\mathbf{Z}_{p}}^{\top} \mathbf{W}_{p} \mathbf{M}\right) \\
&\text { s.t. } \boldsymbol{\alpha}^{\top} \mathbf{1}=1, \sum_{p=1}^{m} \beta_p^2=1, \boldsymbol{\alpha}\geq\mathbf{0}, \boldsymbol{\beta}\geq\mathbf{0}, \\ &{\mathbf{Z}_{p}} {\mathbf{Z}_{p}}^{\top}=\mathbf{I}_{d_p}, {\mathbf{W}_{p}}^{\top} {\mathbf{W}_{p}}=\mathbf{I}_{k},
{\mathbf{M}} {\mathbf{M}}^{\top}=\mathbf{I}_{k},
\end{aligned}
\end{equation}
where $d_p$ is the dimension of $p$-th latent embedding, $\mathbf{M}$ denotes the consensus coefficient matrix. To simplify the optimization process, we impose orthogonality constraint on $\mathbf{Z}_{p}$ rather than $\mathbf{H}_{p}^{(v)}$. 

After attaining $\mathbf{M}$, we conduct $k$-means on it to get the final result.

\subsection{Optimization}
The optimization problem in Eq. \eqref{AWMVC} is not jointly convex when considering all variables simultaneously. Hence, we propose an alternating algorithm to optimize each variable while the others maintain fixed.

\subsubsection{$\mathbf{H}_{p}^{(v)}$ Subproblem}
Considering that the base matrices are independent of each other, we list the optimization process of $p$-th base matrix of $v$-th view $\mathbf{H}_{p}^{(v)}$ as an example. When other variables except for $\mathbf{H}_{p}^{(v)}$ are fixed, the optimization problem in Eq. \eqref{AWMVC} is reformulated as
\begin{equation}\label{wrt_H}
\begin{aligned}
\min \left\|\mathbf{X}^{(v)}-\mathbf{H}_{p}^{(v)} \mathbf{Z}_{p}\right\|_{F}^{2}.
\end{aligned}
\end{equation}

Differentiating the objective function respecting $\mathbf{H}_{p}^{(v)}$ and setting the derivative to zero, it is obtained that $\mathbf{H}_{p}^{(v)}$ is updated by
\begin{equation}\label{opt_H}
\begin{aligned}
\mathbf{H}_{p}^{(v)}=\mathbf{X}^{(v)} {\mathbf{Z}_{p}}^{\top}.
\end{aligned}
\end{equation}

\subsubsection{$\mathbf{M}$ Subproblem}
Fixing $\mathbf{H}$, $\mathbf{Z}$, $\mathbf{W}$, $\boldsymbol{\alpha}$, $\boldsymbol{\beta}$, Eq. \eqref{AWMVC} is reduced to
\begin{equation}\label{opt_M}
\begin{aligned}
\max \operatorname{Tr}\left(\mathbf{M} \mathbf{A}\right) \text { s.t. }
{\mathbf{M}} {\mathbf{M}}^{\top}=\mathbf{I}_{k},
\end{aligned}
\end{equation}
where
\begin{equation}
\begin{aligned}
\mathbf{A}=\sum_{p=1}^{m} \beta_p{\mathbf{Z}_{p}}^{\top} \mathbf{W}_{p}.
\end{aligned}
\end{equation}

Eq. \eqref{opt_M} can be effectively solved via SVD with computational complexity $\mathcal{O}\left(nk^{2}\right)$.

\subsubsection{$\mathbf{W}_{p}$ Subproblem}
With other variables fixed in Eq. \eqref{AWMVC}, $\mathbf{W}_{p}$ can be updated by the following formula
\begin{equation}\label{opt_W}
\begin{aligned}
\max \operatorname{Tr}\left({\mathbf{W}_{p}}^{\top} \mathbf{Z}_{p} {\mathbf{M}^{\top}}\right) \text { s.t. } {\mathbf{W}_{p}}^{\top} {\mathbf{W}_{p}}=\mathbf{I}_{k}.
\end{aligned}
\end{equation}

Similar to Eq. \eqref{opt_M}, it can be effectually solved by SVD with computational complexity $\mathcal{O}\left(d_pk^{2}\right)$.

\subsubsection{$\mathbf{Z}_{p}$ Subproblem}
By dropping the irrelevant variables involved in Eq. \eqref{AWMVC}, the objective formulation concerning $\mathbf{Z}_{p}$ can be rewritten as
\begin{equation}\label{wrt_Z}
\begin{aligned}
&\min \frac{1}{2} \alpha_p^2 \sum_{v=1}^{V} \left\|\mathbf{X}^{(v)}-\mathbf{H}_{p}^{(v)} \mathbf{Z}_{p}\right\|_{F}^{2} - \beta_p\operatorname{Tr}\left({\mathbf{Z}_{p}}^{\top} \mathbf{W}_{p} \mathbf{M}\right) \\
&\text { s.t. } {\mathbf{Z}_{p}} {\mathbf{Z}_{p}}^{\top}=\mathbf{I}_{d_p},
\end{aligned}
\end{equation}
which can be further transformed into
\begin{equation}\label{opt_Z}
\begin{aligned}
\max \operatorname{Tr}\left({\mathbf{Z}_{p}} \mathbf{B}\right)
&\text { s.t. } {\mathbf{Z}_{p}} {\mathbf{Z}_{p}}^{\top}=\mathbf{I}_{d_p},
\end{aligned}
\end{equation}
where
\begin{equation}
\begin{aligned}
\mathbf{B}=\alpha_p^2 \sum_{v=1}^{V}  {\mathbf{X}^{(v)}}^{\top} \mathbf{H}_{p}^{(v)} + \beta_p {\mathbf{M}}^{\top}{\mathbf{W}_{p}}^{\top}.
\end{aligned}
\end{equation}

Same as Eq. \eqref{opt_M}, it can be efficiently solved via SVD with computational complexity $\mathcal{O}\left(n{d_p}^{2}\right)$.

\subsubsection{$\alpha_p$ Subproblem}
Given $\mathbf{H}$, $\mathbf{Z}$, $\mathbf{W}$, $\mathbf{M}$, $\boldsymbol{\beta}$, the formulation respecting $\boldsymbol{\alpha}$ can be solved via optimizing the following formula
\begin{equation}\label{opt_alpha}
\begin{aligned}
\min \sum_{p=1}^{m} \alpha_p^2 r_p^2 \text { s.t. } \boldsymbol{\alpha}^{\top} \mathbf{1}=1, \boldsymbol{\alpha}\geq\mathbf{0},
\end{aligned}
\end{equation}
where
\begin{equation}
\begin{aligned}
r_p^2=\sum_{v=1}^{V} \left\|\mathbf{X}^{(v)}-\mathbf{H}_{p}^{(v)} \mathbf{Z}_{p}\right\|_{F}^{2}.
\end{aligned}
\end{equation}

Based on Cauchy-Schwarz inequality, $\alpha_p$ is updated by
\begin{equation}\label{get_alpha}
\begin{aligned}
\alpha_{p}=\frac{\frac{1}{r_{p}}}{\sum_{p=1}^{m} \frac{1}{r_{p}}}.
\end{aligned}
\end{equation}

\subsubsection{$\beta_p$ Subproblem}
Fixing $\mathbf{H}$, $\mathbf{Z}$, $\mathbf{W}$, $\mathbf{M}$, $\boldsymbol{\alpha}$, Eq. \eqref{AWMVC} is reduced to
\begin{equation}\label{opt_c}
\begin{aligned}
\max \sum_{p=1}^{m} \beta_p\theta_p \text { s.t. } \sum_{p=1}^{m} \beta_p^2=1, \boldsymbol{\beta}\geq\mathbf{0},
\end{aligned}
\end{equation}
where
\begin{equation}
\begin{aligned}
\theta_p=\operatorname{Tr}\left({\mathbf{Z}_{p}}^{\top} \mathbf{W}_{p} \mathbf{M}\right).
\end{aligned}
\end{equation}

The optimal solution for Eq. \eqref{opt_c} is
\begin{equation}\label{get_c}
\begin{aligned}
\beta_p=\frac{\theta_p }{\sqrt{\sum_{p=1}^{m} \theta_p^{2}}}.
\end{aligned}
\end{equation}

\begin{algorithm}[h]
\renewcommand{\algorithmicrequire}{\textbf{Input:}}
	\renewcommand{\algorithmicensure}{\textbf{Output:}}
		\caption{Auto-weighted Multi-view Clustering for Large-scale Data}
		\label{algo}
		\begin{algorithmic}[1]
			\REQUIRE Dataset $\left\{\mathbf{X}^{(v)}\right\}_{v=1}^{V}$, cluster number $k$.
			\ENSURE A consensus coefficient matrix $\mathbf{M}$.
			\STATE Initialize  $\mathbf{Z}$, $\mathbf{W}$, $\boldsymbol{\alpha}=\mathbf{1}/m$, $\boldsymbol{\beta}=\mathbf{1}/\sqrt{m}$.
			\WHILE{not converged}
			\STATE Update $\mathbf{H}_{p}^{(v)}$ via Eq. \eqref{opt_H}.
			\STATE Update $\mathbf{M}$ via Eq. \eqref{opt_M}.
			\STATE Update $\mathbf{W}_p$ via Eq. \eqref{opt_W}.
			\STATE Update $\mathbf{Z}_p$ via Eq. \eqref{opt_Z}.
			\STATE Update $\boldsymbol{\alpha}$ via Eq. \eqref{get_alpha}.
			\STATE Update $\boldsymbol{\beta}$ via Eq. \eqref{get_c}.
			\ENDWHILE
		\end{algorithmic}
	\end{algorithm}
The alternate optimization process of AWMVC is summarized in Algorithm 1. After attaining the consensus coefficient matrix $\mathbf{M}$, we conduct $k$-means on it to obtain the final cluster assignments.

\subsection{Discussion}
Firstly, we analyze the time complexity of our proposed method. Then the convergence of AWMVC is proven theoretically. Afterward, we will discuss the extension of AWMVC.
\subsubsection{Time complexity}
In the optimization process, the time complexity of updating $\mathbf{M}$, $\mathbf{W}_{p}$ and $\mathbf{Z}_{p}$ is provided as $\mathcal{O}\left(n{k}^{2}\right)$, $\mathcal{O}\left(d_pk^{2}\right)$ and $\mathcal{O}\left(n{d_p}^{2}\right)$, respectively. When updating $\mathbf{H}_{p}^{(v)}$, it costs $\mathcal{O}\left({d_v}{d_p}n\right)$ to execute matrix multiplication to obtain the optimal $\mathbf{H}_{p}^{(v)}$. Updating $\boldsymbol{\alpha}$ and $\boldsymbol{\beta}$ takes $\mathcal{O}\left({d_v}{d_p}n\right)$ and $\mathcal{O}\left({k}{d_p}n\right)$ to execute matrix multiplication, then costs $\mathcal{O}\left(m\right)$ to calculate them, separately. Thus, 
at each iteration, the time complexity of AWMVC is $\mathcal{O}\left(n{k}^{2}+\sum_{p=1}^{m}\left(d_pk^{2}+n{d_p}^{2}\right)+\sum_{v=1}^{V}\sum_{p=1}^{m}{d_v}{d_p}n\right)$, which is linear to the data number $n$. In addition, our proposed method is free of hyperparameters. Therefore, it is suitable for large-scale data.
\subsubsection{Convergence}
The objective value in Eq. \eqref{AWMVC} decreases monotonically when one variable updates with the others fixed. Therefore, to prove the convergence of AWMVC, we merely need to show that the formulation has a lower bound. The proven process is provided as follows: By Cauchy-Schwartz inequality, we have
\begin{equation}\label{Cauchy-Schwartz_1}
\begin{aligned}
& \operatorname{Tr}\left({\mathbf{Z}_{p}}^{\top} \mathbf{W}_{p} \mathbf{M}\right)
\le\Vert {\mathbf{Z}}_{p}^{\top} \Vert_F \Vert {\mathbf{W}_{p}}\Vert_F \Vert\mathbf{M} \Vert_F.
\end{aligned}
\end{equation}

Considering that $\boldsymbol{\beta}$ satisfies $\sum_{p=1}^{m} \beta_p^2=1$ and $\boldsymbol{\beta}\geq0$, it is easy to conclude that $\boldsymbol{\beta}\leq\mathbf{1}$. Based on the above analysis, we can obtain
\begin{equation}\label{higher_bound}
\begin{aligned}
&\sum_{p=1}^{m} \beta_p\operatorname{Tr}\left({\mathbf{Z}_{p}}^{\top} \mathbf{W}_{p} \mathbf{M}\right)\le
\sum_{p=1}^{m} \operatorname{Tr}\left({\mathbf{Z}_{p}}^{\top} \mathbf{W}_{p} \mathbf{M}\right)
\\&\le
\sum_{p=1}^{m}\Vert {\mathbf{Z}}_{p}^{\top} \Vert_F \Vert {\mathbf{W}_{p}}\Vert_F \Vert\mathbf{M} \Vert_F=\sum_{p=1}^{m} k\sqrt{d_p}.
\end{aligned}
\end{equation}

Therefore, there is a lower bound for the formula of Eq. \eqref{AWMVC}, with a specific value as follows
\begin{equation}
\begin{aligned}
& \sum_{p=1}^{m} \sum_{v=1}^{V}\frac{1}{2}\alpha_p^2 \left\|\mathbf{X}^{(v)}-\mathbf{H}_{p}^{(v)} \mathbf{Z}_{p}\right\|_{F}^{2} -\sum_{p=1}^{m} \beta_p\operatorname{Tr}\left({\mathbf{Z}_{p}}^{\top} \mathbf{W}_{p} \mathbf{M}\right)
\\& \geq 0-\sum_{p=1}^{m} k\sqrt{d_p}=-\sum_{p=1}^{m} k\sqrt{d_p}.
\end{aligned}
\end{equation}

In our alternating optimization process, the objective value monotonically decreases with each iteration. As a consequence, the algorithm is convergent in theoretic. Furthermore, we will verify the convergence of AWMVC in our experiment.

\subsubsection{Extension}
Our proposed method gives the inspiration to map data features into different latent embedding under specific dimensions and fuse the embedding efficiently. This way, the consensus matrix extracts more complementary and comprehensive information than a fixed dimension. In addition, the model is free of hyperparameters and fitting to handle large-scale datasets. Also, this idea can easily extend to various fields, including multi-view clustering. 
\section{Experimental Results}
In this section, we conduct comprehensive experiments on various large-scale benchmark datasets to verify the excellent performance of AWMVC, involving clustering performance comparison, convergence and evolution, and running time comparison. Moreover, we analyze dimension weights and conduct an ablation study to evaluate the validity of our proposed algorithm. 
\subsection{Experiment Settings}
Seven benchmark datasets are adopted to verify the promising performance of AWMVC, and the maximum number of samples used is more than 100,000. The datasets used in our experiment include Flower17\footnote{\url{https://www.robots.ox.ac.uk/∼vgg/data/flowers/}}, AwA\footnote{https://cvml.ist.ac.at/AwA/}, Caltech256\footnote{\url{https://www.kaggle.com/datasets/jessicali9530/caltech256/}}, MNIST\footnote{\url{http://yann.lecun.com/exdb/mnist/}}, VGGFace2\footnote{\url{http://www.robots.ox.ac.uk/~vgg/data/vgg_face2/}}, TinyImageNet\footnote{\url{http://cs231n.stanford.edu/tiny-imagenet-200.zip}}, YouTubeFace50\footnote{\url{http://archive.ics.uci.edu/ml/datasets/YouTube+Multiview+Video+Games+Dataset}}. The detailed information of each dataset is listed in Table \ref{dataset}.
 \begin{table}
 \begin{center}
 \caption{Datasets used in our experiments.} 
 \label{dataset}
\begin{tabular}{ |c| c| c| c|}
\hline
		Datasets  		&Samples   &Views       &Clusters    \\
		\hline
		Flower17     	    &1360      &7           &17  		 \tabularnewline \hline 
		AwA		    &30475	   &6      	&50		 \tabularnewline \hline 
		Caltech256	    	&30607      &4          &257		 \tabularnewline \hline 
		MNIST     	    &60000      &3           &10  		 \tabularnewline \hline 
		VGGFace2	    	&72283      &4           &200 		\tabularnewline \hline 
		TinyImageNet			&100000      &4           &200 		 \tabularnewline \hline 
		YouTubeFace50     &126054       &4          &50   \tabularnewline \hline 
\end{tabular}
\end{center}
\end{table}

\begin{table*}[h]
	\centering
	\caption{Empirical evaluation and comparison of AWMVC with eight baseline methods on $7$  benchmark datasets in terms of clustering accuracy (ACC), normalized mutual information (NMI), Purity, and Fscore.}
	\label{result}
	\small 
	\begin{tabular}{|c|c|c|c|c|c|c|c|c|c|c|}
		\hline 
\multicolumn{1}{|c|}{{\color[HTML]{000000} Datasets}}              & \multicolumn{1}{c|}{{\color[HTML]{000000} AMGL}}  & \multicolumn{1}{c|}{{\color[HTML]{000000} UOMVSC}} & \multicolumn{1}{c|}{{\color[HTML]{000000} MNMF}}  & \multicolumn{1}{c|}{{\color[HTML]{000000} SMVSC}}      & \multicolumn{1}{c|}{{\color[HTML]{000000} BMVC}}        & \multicolumn{1}{c|}{{\color[HTML]{000000} LMVSC}}       & \multicolumn{1}{c|}{{\color[HTML]{000000} FPMVS-CAG}}      & \multicolumn{1}{c|}{{\color[HTML]{000000} OPMC}}           & {\color[HTML]{000000} AWMVC}          \\ \hline
\multicolumn{1}{|c|}{{\color[HTML]{000000} Hyperparameter number}} & \multicolumn{1}{c|}{{\color[HTML]{000000} 0}}     & \multicolumn{1}{c|}{{\color[HTML]{000000} 1}}      & \multicolumn{1}{c|}{{\color[HTML]{000000} 1}}     & \multicolumn{1}{c|}{{\color[HTML]{000000} 1}}          & \multicolumn{1}{c|}{{\color[HTML]{000000} 4}}           & \multicolumn{1}{c|}{{\color[HTML]{000000} 1}}           & \multicolumn{1}{c|}{{\color[HTML]{000000} 0}}              & \multicolumn{1}{c|}{{\color[HTML]{000000} 0}}              & {\color[HTML]{000000} 0}              \\ \hline
\multicolumn{10}{|c|}{{\color[HTML]{000000} ACC}}                                                                                                               \\ \hline
\multicolumn{1}{|c|}{{\color[HTML]{000000} Flower17}}              & \multicolumn{1}{c|}{{\color[HTML]{000000} 9.70}}  & \multicolumn{1}{c|}{{\color[HTML]{000000} 36.32}}  & \multicolumn{1}{c|}{{\color[HTML]{000000} 35.81}} & \multicolumn{1}{c|}{{\color[HTML]{000000} 27.94}}      & \multicolumn{1}{c|}{{\color[HTML]{000000} 26.99}}       & \multicolumn{1}{c|}{{\color[HTML]{000000} {\ul 40.37}}} & \multicolumn{1}{c|}{{\color[HTML]{000000} 26.03}}          & \multicolumn{1}{c|}{{\color[HTML]{000000} 32.13}}          & {\color[HTML]{000000} \textbf{49.12}} \\ \hline
\multicolumn{1}{|c|}{{\color[HTML]{000000} AwA}}                   & \multicolumn{1}{c|}{{\color[HTML]{000000} -}}     & \multicolumn{1}{c|}{{\color[HTML]{000000} -}}      & \multicolumn{1}{c|}{{\color[HTML]{000000} 7.72}}  & \multicolumn{1}{c|}{{\color[HTML]{000000} { 9.19}}} & \multicolumn{1}{c|}{{\color[HTML]{000000} 8.67}}        & \multicolumn{1}{c|}{{\color[HTML]{000000} 7.97}}        & \multicolumn{1}{c|}{{\color[HTML]{000000} 9.11}}           & \multicolumn{1}{c|}{{\color[HTML]{000000} \ul9.26}}           & {\color[HTML]{000000} \textbf{9.42}}  \\ \hline
\multicolumn{1}{|c|}{{\color[HTML]{000000} Caltech256}}            & \multicolumn{1}{c|}{{\color[HTML]{000000} -}}     & \multicolumn{1}{c|}{{\color[HTML]{000000} -}}      & \multicolumn{1}{c|}{{\color[HTML]{000000} 2.71}}  & \multicolumn{1}{c|}{{\color[HTML]{000000} 9.67}}       & \multicolumn{1}{c|}{{\color[HTML]{000000} 8.63}}        & \multicolumn{1}{c|}{{\color[HTML]{000000} 9.87}}        & \multicolumn{1}{c|}{{\color[HTML]{000000} 9.49}}           & \multicolumn{1}{c|}{{\color[HTML]{000000} {\ul 11.20}}}    & {\color[HTML]{000000} \textbf{12.75}} \\ \hline
\multicolumn{1}{|c|}{{\color[HTML]{000000} MNIST}}                 & \multicolumn{1}{c|}{{\color[HTML]{000000} -}}     & \multicolumn{1}{c|}{{\color[HTML]{000000} -}}      & \multicolumn{1}{c|}{{\color[HTML]{000000} 98.06}} & \multicolumn{1}{c|}{{\color[HTML]{000000} 98.75}}      & \multicolumn{1}{c|}{{\color[HTML]{000000} 45.95}}       & \multicolumn{1}{c|}{{\color[HTML]{000000} { 98.57}}} & \multicolumn{1}{c|}{{\color[HTML]{000000} \ul98.84}}          & \multicolumn{1}{c|}{{\color[HTML]{000000} 98.58}}          & {\color[HTML]{000000} \textbf{98.85}} \\ \hline
\multicolumn{1}{|c|}{{\color[HTML]{000000} VGGFace2}}              & \multicolumn{1}{c|}{{\color[HTML]{000000} -}}     & \multicolumn{1}{c|}{{\color[HTML]{000000} -}}      & \multicolumn{1}{c|}{{\color[HTML]{000000} 0.90}}  & \multicolumn{1}{c|}{{\color[HTML]{000000} 3.08}}       & \multicolumn{1}{c|}{{\color[HTML]{000000} 3.99}}        & \multicolumn{1}{c|}{{\color[HTML]{000000} {\ul 4.40}}}  & \multicolumn{1}{c|}{{\color[HTML]{000000} 3.15}}           & \multicolumn{1}{c|}{{\color[HTML]{000000} 3.88}}           & {\color[HTML]{000000} \textbf{6.84}}  \\ \hline
\multicolumn{1}{|c|}{{\color[HTML]{000000} TinyImageNet}}          & \multicolumn{1}{c|}{{\color[HTML]{000000} -}}     & \multicolumn{1}{c|}{{\color[HTML]{000000} -}}      & \multicolumn{1}{c|}{{\color[HTML]{000000} 0.50}}  & \multicolumn{1}{c|}{{\color[HTML]{000000} 3.03}}       & \multicolumn{1}{c|}{{\color[HTML]{000000} 4.09}}        & \multicolumn{1}{c|}{{\color[HTML]{000000} { 4.39}}}  & \multicolumn{1}{c|}{{\color[HTML]{000000} 2.93}}           & \multicolumn{1}{c|}{{\color[HTML]{000000} \ul5.15}}           & {\color[HTML]{000000} \textbf{5.17}}  \\ \hline
\multicolumn{1}{|c|}{{\color[HTML]{000000} YouTubeFace50}}         & \multicolumn{1}{c|}{{\color[HTML]{000000} -}}     & \multicolumn{1}{c|}{{\color[HTML]{000000} -}}      & \multicolumn{1}{c|}{{\color[HTML]{000000} 4.82}}  & \multicolumn{1}{c|}{{\color[HTML]{000000} 67.79}}      & \multicolumn{1}{c|}{{\color[HTML]{000000} 66.00}}       & \multicolumn{1}{c|}{{\color[HTML]{000000} {\ul 72.73}}} & \multicolumn{1}{c|}{{\color[HTML]{000000} 66.31}}          & \multicolumn{1}{c|}{{\color[HTML]{000000} 69.36}}          & {\color[HTML]{000000} \textbf{75.54}} \\ \hline
\multicolumn{10}{|c|}{{\color[HTML]{000000} NMI}}                                                               
\\ \hline
\multicolumn{1}{|c|}{{\color[HTML]{000000} Flower17}}              & \multicolumn{1}{c|}{{\color[HTML]{000000} 10.25}} & \multicolumn{1}{c|}{{\color[HTML]{000000} 34.95}}  & \multicolumn{1}{c|}{{\color[HTML]{000000} 35.25}} & \multicolumn{1}{c|}{{\color[HTML]{000000} 25.12}}      & \multicolumn{1}{c|}{{\color[HTML]{000000} 25.62}}       & \multicolumn{1}{c|}{{\color[HTML]{000000} {\ul 38.29}}} & \multicolumn{1}{c|}{{\color[HTML]{000000} 26.04}}          & \multicolumn{1}{c|}{{\color[HTML]{000000} 29.69}}          & {\color[HTML]{000000} \textbf{49.78}} \\ \hline
\multicolumn{1}{|c|}{{\color[HTML]{000000} AwA}}                   & \multicolumn{1}{c|}{{\color[HTML]{000000} -}}     & \multicolumn{1}{c|}{{\color[HTML]{000000} -}}      & \multicolumn{1}{c|}{{\color[HTML]{000000} 9.65}}  & \multicolumn{1}{c|}{{\color[HTML]{000000} 10.68}}      & \multicolumn{1}{c|}{{\color[HTML]{000000} {\ul 11.95}}} & \multicolumn{1}{c|}{{\color[HTML]{000000} 9.32}}        & \multicolumn{1}{c|}{{\color[HTML]{000000} 10.84}}          & \multicolumn{1}{c|}{{\color[HTML]{000000} \textbf{12.22}}} & {\color[HTML]{000000} 11.31}          \\ \hline
\multicolumn{1}{|c|}{{\color[HTML]{000000} Caltech256}}            & \multicolumn{1}{c|}{{\color[HTML]{000000} -}}     & \multicolumn{1}{c|}{{\color[HTML]{000000} -}}      & \multicolumn{1}{c|}{{\color[HTML]{000000} 1.96}}  & \multicolumn{1}{c|}{{\color[HTML]{000000} 24.42}}      & \multicolumn{1}{c|}{{\color[HTML]{000000} 31.83}}       & \multicolumn{1}{c|}{{\color[HTML]{000000} 32.14}}       & \multicolumn{1}{c|}{{\color[HTML]{000000} 22.03}}          & \multicolumn{1}{c|}{{\color[HTML]{000000} {\ul 32.72}}}    & {\color[HTML]{000000} \textbf{34.61}} \\ \hline
\multicolumn{1}{|c|}{{\color[HTML]{000000} MNIST}}                 & \multicolumn{1}{c|}{{\color[HTML]{000000} -}}     & \multicolumn{1}{c|}{{\color[HTML]{000000} -}}      & \multicolumn{1}{c|}{{\color[HTML]{000000} 94.74}} & \multicolumn{1}{c|}{{\color[HTML]{000000} 96.27}}      & \multicolumn{1}{c|}{{\color[HTML]{000000} 39.59}}       & \multicolumn{1}{c|}{{\color[HTML]{000000} 95.93}}       & \multicolumn{1}{c|}{{\color[HTML]{000000} \textbf{96.51}}} & \multicolumn{1}{c|}{{\color[HTML]{000000} 95.85}}          & {\color[HTML]{000000} {\ul 96.47}}    \\ \hline
\multicolumn{1}{|c|}{{\color[HTML]{000000} VGGFace2}}              & \multicolumn{1}{c|}{{\color[HTML]{000000} -}}     & \multicolumn{1}{c|}{{\color[HTML]{000000} -}}      & \multicolumn{1}{c|}{{\color[HTML]{000000} 0.61}}  & \multicolumn{1}{c|}{{\color[HTML]{000000} 10.35}}      & \multicolumn{1}{c|}{{\color[HTML]{000000} {\ul 15.04}}} & \multicolumn{1}{c|}{{\color[HTML]{000000} 14.04}}       & \multicolumn{1}{c|}{{\color[HTML]{000000} 9.57}}           & \multicolumn{1}{c|}{{\color[HTML]{000000} 13.29}}          & {\color[HTML]{000000} \textbf{18.59}} \\ \hline
\multicolumn{1}{|c|}{{\color[HTML]{000000} TinyImageNet}}          & \multicolumn{1}{c|}{{\color[HTML]{000000} -}}     & \multicolumn{1}{c|}{{\color[HTML]{000000} -}}      & \multicolumn{1}{c|}{{\color[HTML]{000000} 0.41}}  & \multicolumn{1}{c|}{{\color[HTML]{000000} 10.64}}      & \multicolumn{1}{c|}{{\color[HTML]{000000} 13.75}}       & \multicolumn{1}{c|}{{\color[HTML]{000000} 13.32}}       & \multicolumn{1}{c|}{{\color[HTML]{000000} 10.16}}          & \multicolumn{1}{c|}{{\color[HTML]{000000} \textbf{16.13}}} & {\color[HTML]{000000} {\ul 14.54}}    \\ \hline
\multicolumn{1}{|c|}{{\color[HTML]{000000} YouTubeFace50}}         & \multicolumn{1}{c|}{{\color[HTML]{000000} -}}     & \multicolumn{1}{c|}{{\color[HTML]{000000} -}}      & \multicolumn{1}{c|}{{\color[HTML]{000000} 0.08}}  & \multicolumn{1}{c|}{{\color[HTML]{000000} 82.76}}      & \multicolumn{1}{c|}{{\color[HTML]{000000} 81.90}}       & \multicolumn{1}{c|}{{\color[HTML]{000000} {\ul 83.98}}} & \multicolumn{1}{c|}{{\color[HTML]{000000} 83.51}}          & \multicolumn{1}{c|}{{\color[HTML]{000000} 82.36}}          & {\color[HTML]{000000} \textbf{85.97}} \\ \hline
\multicolumn{10}{|c|}{{\color[HTML]{000000} Purity}}                                                                           \\ \hline
\multicolumn{1}{|c|}{{\color[HTML]{000000} Flower17}}              & \multicolumn{1}{c|}{{\color[HTML]{000000} 10.76}} & \multicolumn{1}{c|}{{\color[HTML]{000000} 37.94}}  & \multicolumn{1}{c|}{{\color[HTML]{000000} 37.06}} & \multicolumn{1}{c|}{{\color[HTML]{000000} 29.26}}      & \multicolumn{1}{c|}{{\color[HTML]{000000} 29.41}}       & \multicolumn{1}{c|}{{\color[HTML]{000000} {\ul 41.69}}} & \multicolumn{1}{c|}{{\color[HTML]{000000} 27.43}}          & \multicolumn{1}{c|}{{\color[HTML]{000000} 33.60}}          & {\color[HTML]{000000} \textbf{51.18}} \\ \hline
\multicolumn{1}{|c|}{{\color[HTML]{000000} AwA}}                   & \multicolumn{1}{c|}{{\color[HTML]{000000} -}}     & \multicolumn{1}{c|}{{\color[HTML]{000000} -}}      & \multicolumn{1}{c|}{{\color[HTML]{000000} 9.99}}  & \multicolumn{1}{c|}{{\color[HTML]{000000} 10.02}}      & \multicolumn{1}{c|}{{\color[HTML]{000000} 10.94}}       & \multicolumn{1}{c|}{{\color[HTML]{000000} 10.25}}       & \multicolumn{1}{c|}{{\color[HTML]{000000} 9.69}}           & \multicolumn{1}{c|}{{\color[HTML]{000000} {\ul 11.19}}}    & {\color[HTML]{000000} \textbf{11.51}} \\ \hline
\multicolumn{1}{|c|}{{\color[HTML]{000000} Caltech256}}            & \multicolumn{1}{c|}{{\color[HTML]{000000} -}}     & \multicolumn{1}{c|}{{\color[HTML]{000000} -}}      & \multicolumn{1}{c|}{{\color[HTML]{000000} 3.54}}  & \multicolumn{1}{c|}{{\color[HTML]{000000} 11.42}}      & \multicolumn{1}{c|}{{\color[HTML]{000000} 14.94}}       & \multicolumn{1}{c|}{{\color[HTML]{000000} 16.52}}       & \multicolumn{1}{c|}{{\color[HTML]{000000} 11.07}}          & \multicolumn{1}{c|}{{\color[HTML]{000000} {\ul 16.93}}}    & {\color[HTML]{000000} \textbf{18.83}} \\ \hline
\multicolumn{1}{|c|}{{\color[HTML]{000000} MNIST}}                 & \multicolumn{1}{c|}{{\color[HTML]{000000} -}}     & \multicolumn{1}{c|}{{\color[HTML]{000000} -}}      & \multicolumn{1}{c|}{{\color[HTML]{000000} 98.06}} & \multicolumn{1}{c|}{{\color[HTML]{000000} 98.75}}      & \multicolumn{1}{c|}{{\color[HTML]{000000} 47.66}}       & \multicolumn{1}{c|}{{\color[HTML]{000000} 98.57}}       & \multicolumn{1}{c|}{{\color[HTML]{000000} {\ul 98.84}}}    & \multicolumn{1}{c|}{{\color[HTML]{000000} 98.58}}          & {\color[HTML]{000000} \textbf{98.85}} \\ \hline
\multicolumn{1}{|c|}{{\color[HTML]{000000} VGGFace2}}              & \multicolumn{1}{c|}{{\color[HTML]{000000} -}}     & \multicolumn{1}{c|}{{\color[HTML]{000000} -}}      & \multicolumn{1}{c|}{{\color[HTML]{000000} 1.17}}  & \multicolumn{1}{c|}{{\color[HTML]{000000} 3.15}}       & \multicolumn{1}{c|}{{\color[HTML]{000000} 4.66}}        & \multicolumn{1}{c|}{{\color[HTML]{000000} {\ul 5.05}}}  & \multicolumn{1}{c|}{{\color[HTML]{000000} 3.22}}           & \multicolumn{1}{c|}{{\color[HTML]{000000} 4.43}}           & {\color[HTML]{000000} \textbf{7.57}}  \\ \hline
\multicolumn{1}{|c|}{{\color[HTML]{000000} TinyImageNet}}          & \multicolumn{1}{c|}{{\color[HTML]{000000} -}}     & \multicolumn{1}{c|}{{\color[HTML]{000000} -}}      & \multicolumn{1}{c|}{{\color[HTML]{000000} 0.70}}  & \multicolumn{1}{c|}{{\color[HTML]{000000} 3.18}}       & \multicolumn{1}{c|}{{\color[HTML]{000000} 4.69}}        & \multicolumn{1}{c|}{{\color[HTML]{000000} 5.05}}        & \multicolumn{1}{c|}{{\color[HTML]{000000} 2.96}}           & \multicolumn{1}{c|}{{\color[HTML]{000000} \textbf{5.88}}}  & {\color[HTML]{000000} {\ul 5.77}}     \\ \hline
\multicolumn{1}{|c|}{{\color[HTML]{000000} YouTubeFace50}}         & \multicolumn{1}{c|}{{\color[HTML]{000000} -}}     & \multicolumn{1}{c|}{{\color[HTML]{000000} -}}      & \multicolumn{1}{c|}{{\color[HTML]{000000} 4.85}}  & \multicolumn{1}{c|}{{\color[HTML]{000000} 71.09}}      & \multicolumn{1}{c|}{{\color[HTML]{000000} 73.64}}       & \multicolumn{1}{c|}{{\color[HTML]{000000} {\ul 78.24}}} & \multicolumn{1}{c|}{{\color[HTML]{000000} 69.32}}          & \multicolumn{1}{c|}{{\color[HTML]{000000} 72.53}}          & {\color[HTML]{000000} \textbf{79.31}} \\ \hline
\multicolumn{10}{|c|}{{\color[HTML]{000000} Fscore}}                                                                                                                                                                                                                                                                                                                                                                                                                                                                                                                           \\ \hline
\multicolumn{1}{|c|}{{\color[HTML]{000000} Flower17}}              & \multicolumn{1}{c|}{{\color[HTML]{000000} 11.49}} & \multicolumn{1}{c|}{{\color[HTML]{000000} 19.69}}  & \multicolumn{1}{c|}{{\color[HTML]{000000} 23.00}} & \multicolumn{1}{c|}{{\color[HTML]{000000} 16.00}}      & \multicolumn{1}{c|}{{\color[HTML]{000000} 16.61}}       & \multicolumn{1}{c|}{{\color[HTML]{000000} {\ul 26.12}}} & \multicolumn{1}{c|}{{\color[HTML]{000000} 16.65}}          & \multicolumn{1}{c|}{{\color[HTML]{000000} 12.85}}          & {\color[HTML]{000000} \textbf{34.60}} \\ \hline
\multicolumn{1}{|c|}{{\color[HTML]{000000} AwA}}                   & \multicolumn{1}{c|}{{\color[HTML]{000000} -}}     & \multicolumn{1}{c|}{{\color[HTML]{000000} -}}      & \multicolumn{1}{c|}{{\color[HTML]{000000} 4.15}}  & \multicolumn{1}{c|}{{\color[HTML]{000000} {\ul 6.15}}} & \multicolumn{1}{c|}{{\color[HTML]{000000} 4.32}}        & \multicolumn{1}{c|}{{\color[HTML]{000000} 3.99}}        & \multicolumn{1}{c|}{{\color[HTML]{000000} \textbf{6.40}}}  & \multicolumn{1}{c|}{{\color[HTML]{000000} 2.47}}           & {\color[HTML]{000000} 4.56}           \\ \hline
\multicolumn{1}{|c|}{{\color[HTML]{000000} Caltech256}}            & \multicolumn{1}{c|}{{\color[HTML]{000000} -}}     & \multicolumn{1}{c|}{{\color[HTML]{000000} -}}      & \multicolumn{1}{c|}{{\color[HTML]{000000} 1.18}}  & \multicolumn{1}{c|}{{\color[HTML]{000000} 5.09}}       & \multicolumn{1}{c|}{{\color[HTML]{000000} 6.25}}        & \multicolumn{1}{c|}{{\color[HTML]{000000} 6.54}}        & \multicolumn{1}{c|}{{\color[HTML]{000000} 5.65}}           & \multicolumn{1}{c|}{{\color[HTML]{000000} {\ul 9.26}}}     & {\color[HTML]{000000} \textbf{11.03}} \\ \hline
\multicolumn{1}{|c|}{{\color[HTML]{000000} MNIST}}                 & \multicolumn{1}{c|}{{\color[HTML]{000000} -}}     & \multicolumn{1}{c|}{{\color[HTML]{000000} -}}      & \multicolumn{1}{c|}{{\color[HTML]{000000} 96.19}} & \multicolumn{1}{c|}{{\color[HTML]{000000} 97.51}}      & \multicolumn{1}{c|}{{\color[HTML]{000000} 33.57}}       & \multicolumn{1}{c|}{{\color[HTML]{000000} 97.19}}       & \multicolumn{1}{c|}{{\color[HTML]{000000} {\ul 97.68}}}    & \multicolumn{1}{c|}{{\color[HTML]{000000} 96.85}}          & {\color[HTML]{000000} \textbf{97.71}} \\ \hline
\multicolumn{1}{|c|}{{\color[HTML]{000000} VGGFace2}}              & \multicolumn{1}{c|}{{\color[HTML]{000000} -}}     & \multicolumn{1}{c|}{{\color[HTML]{000000} -}}      & \multicolumn{1}{c|}{{\color[HTML]{000000} 1.06}}  & \multicolumn{1}{c|}{{\color[HTML]{000000} 1.46}}       & \multicolumn{1}{c|}{{\color[HTML]{000000} 1.46}}        & \multicolumn{1}{c|}{{\color[HTML]{000000} 1.43}}        & \multicolumn{1}{c|}{{\color[HTML]{000000} {\ul 1.46}}}     & \multicolumn{1}{c|}{{\color[HTML]{000000} 0.72}}           & {\color[HTML]{000000} \textbf{2.43}}  \\ \hline
\multicolumn{1}{|c|}{{\color[HTML]{000000} TinyImageNet}}          & \multicolumn{1}{c|}{{\color[HTML]{000000} -}}     & \multicolumn{1}{c|}{{\color[HTML]{000000} -}}      & \multicolumn{1}{c|}{{\color[HTML]{000000} 0.99}}  & \multicolumn{1}{c|}{{\color[HTML]{000000} 1.53}}       & \multicolumn{1}{c|}{{\color[HTML]{000000} 1.55}}        & \multicolumn{1}{c|}{{\color[HTML]{000000} 1.38}}        & \multicolumn{1}{c|}{{\color[HTML]{000000} \textbf{1.75}}}  & \multicolumn{1}{c|}{{\color[HTML]{000000} 1.30}}           & {\color[HTML]{000000} {\ul 1.70}}     \\ \hline
\multicolumn{1}{|c|}{{\color[HTML]{000000} YouTubeFace50}}         & \multicolumn{1}{c|}{{\color[HTML]{000000} -}}     & \multicolumn{1}{c|}{{\color[HTML]{000000} -}}      & \multicolumn{1}{c|}{{\color[HTML]{000000} 4.39}}  & \multicolumn{1}{c|}{{\color[HTML]{000000} 60.74}}      & \multicolumn{1}{c|}{{\color[HTML]{000000} 57.09}}       & \multicolumn{1}{c|}{{\color[HTML]{000000} {\ul 67.14}}} & \multicolumn{1}{c|}{{\color[HTML]{000000} 61.99}}          & \multicolumn{1}{c|}{{\color[HTML]{000000} 62.06}}          & {\color[HTML]{000000} \textbf{70.49}} \\ \hline
	\end{tabular}

\end{table*}
Our proposed algorithm is compared with eight state-of-the-art methods, and the comparative algorithms are summarized as follows
\begin{enumerate}
\item \textbf{Parameter-free auto-weighted multiple graph learning (AMGL)} \cite{10.5555/3060832.3060884}. The algorithm automatically learns an optimal weight for each graph and obtains an optimal global result.
\item \textbf{Unified One-step Multi-view Spectral Clustering (UOMVSC)} \cite{9769920}. This work integrates spectral embedding and $k$-means into a unified framework.
\item \textbf{Multi-view clustering via joint nonnegative matrix factorization (MNMF)} \cite{liu2013multi}. MNMF proposes a matrix-factorization framework to push clustering results of each view into a consensus one.
\item \textbf{Scalable Multi-view Subspace Clustering with Unified Anchors (SMVSC)} \cite{Sun2021ScalableMS}. SMVSC obtains consensus anchor points to get more discriminative information and puts anchor learning and subspace learning into a unified optimization framework.
\item \textbf{Binary Multi-View Clustering (BMVC)} \cite{8387526}. This paper puts discrete representation learning and binary clustering structure learning together.
\item \textbf{Large-scale Multi-view Subspace Clustering in Linear Time (LMVSC)} \cite{Kang2020LargescaleMS}. LMVSC integrates anchor graphs and conducts spectral clustering on a smaller graph.
\item \textbf{Fast Parameter-Free Multi-View Subspace Clustering With Consensus Anchor Guidance (FPMVS-CAG)} \cite{9646486}. This paper simultaneously conducts anchor selection and subspace graph construction and proposes a parameter-free algorithm.
\item \textbf{One-pass Multi-view Clustering for Large-scale Data (OPMC)} \cite{9710821}. OPMC removes the non-negativity constraint of NMFMVC and proposes a method to obtain a discrete clustering partition matrix.
\end{enumerate}
The implementations of the above methods are public on the corresponding papers, and we run them without any changes. Considering that all the methods need to conduct k-means to attain final cluster assignments, we run 50 times k-means to eliminate the randomness in initialization. For methods with hyperparameters, we tune them with grid-search recommended in their papers and report the best results. In our experiment, we simply set $m=3$ and the dimensions range from $k$ to $mk$. All the experiments are conducted on a desktop computer with Intel(R) Core(TM) i9-10850K CPU and 96G RAM.

\subsection{Experiment Results}
We compare AWMVC with eight algorithms on seven widely used benchmark datasets based on four clustering metrics, including accuracy (ACC), normalized mutual information (NMI), purity, and Fscore. The results are shown in Table \ref{result}. It is worth noting that the best outcome is marked in bold, and the second best is underlined. In addition, '\--{}' indicates that the algorithm fails to run smoothly due to an out-of-memory error. From the table, it is observed that 
\begin{enumerate}
\item The proposed method consistently exceeds the competitors on Flower17, Caltech256, VGGFace2, and YouTubeFace50. On other datasets, AWMVC also demonstrates comparable results. For instance, AWMVC outperforms the second best algorithm on all datasets by 21.67\%, 1.73\%, 13.84\%, 0.01\%, 55.45\%, 0.39\%, and 3.86\% in terms of ACC, respectively. The improvements on others clustering indicators are similar. The superior clustering performance verifies the superiority of AWMVC.
\item Most large-scale algorithms fail to handle regular datasets, and AWMVC overcomes this drawback. For instance, anchor-based methods always utilize anchor points to represent the data cloud. However, cause of the information loss between anchors and the data cloud, this method is hard to work well on regular datasets. AWMVC is suitable for diverse applications. Our approach attains promising results for routine datasets like Flower17 and Caltech256, and on large-scale scenarios such as YouTubeFace50, the results are the same excellent.
\item Compared with SMVSC, BMVC, and LMVSC, which suffer from hyperparameters, AWMVC achieves better results. In practical applications, these methods ought to tune the hyperparameters many times to choose the best one, resulting in huge time and space resource consumption. In contrast, our proposed method is parameter-free and more suitable for practical situations. 
\end{enumerate}
 
 \subsection{Convergence and Evolution}
 \begin{figure}[htbp]
	\centering
	
	\subfigure{
		\includegraphics[width=0.23\textwidth]{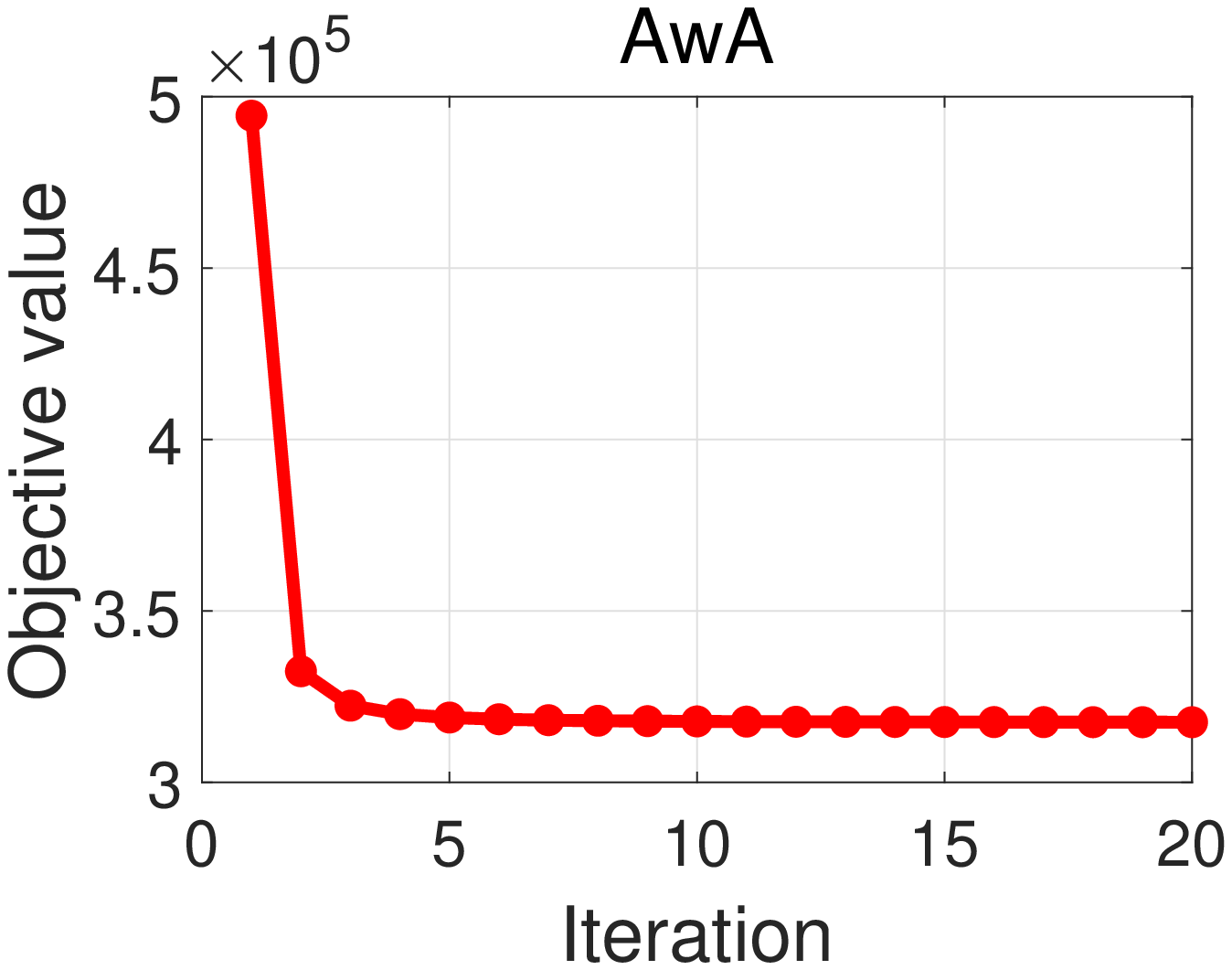}}
	\hspace{-0.2cm}
	\subfigure{
		\includegraphics[width=0.23\textwidth]{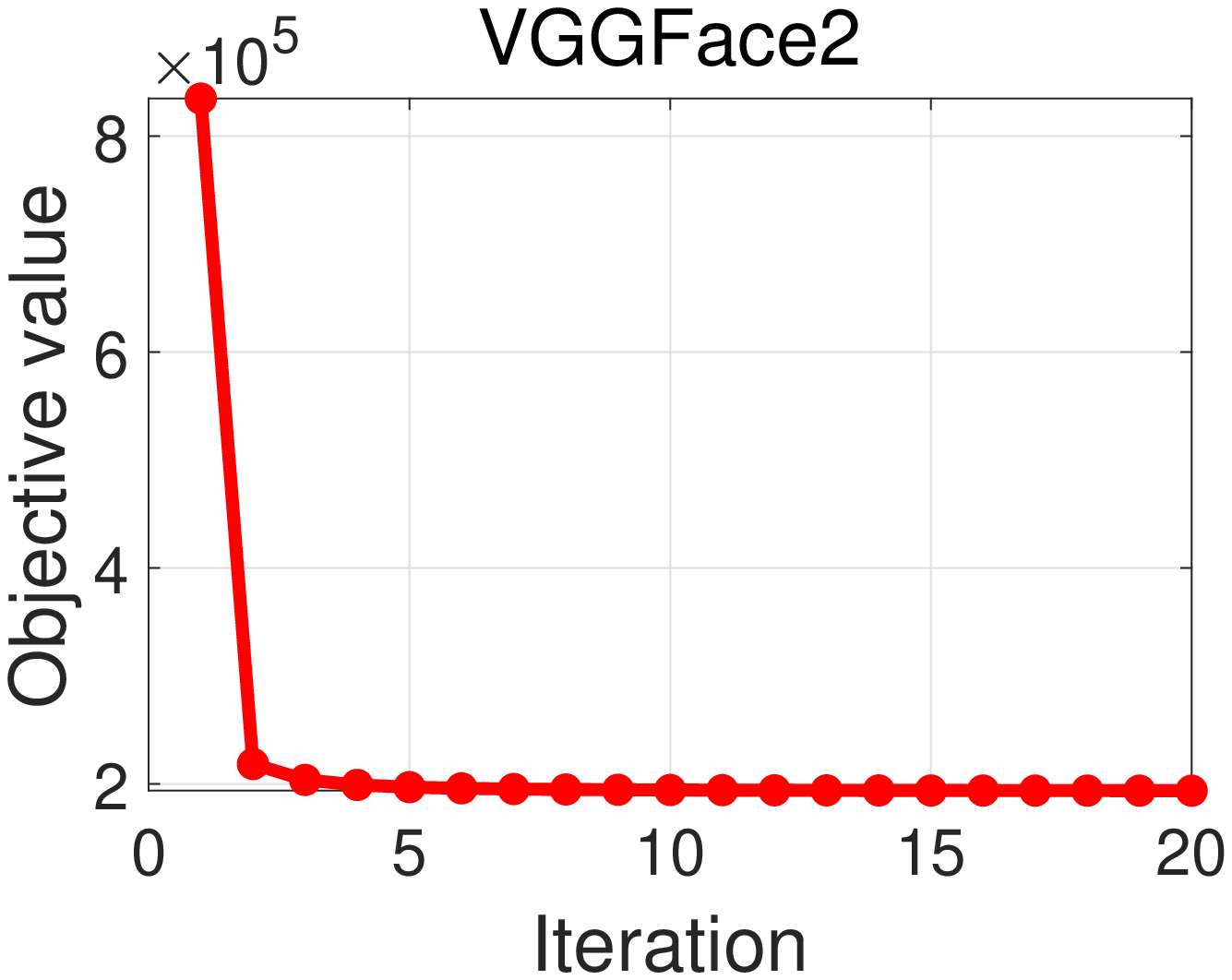}}
\hspace{-0.2cm}
	\subfigure{
		\includegraphics[width=0.23\textwidth]{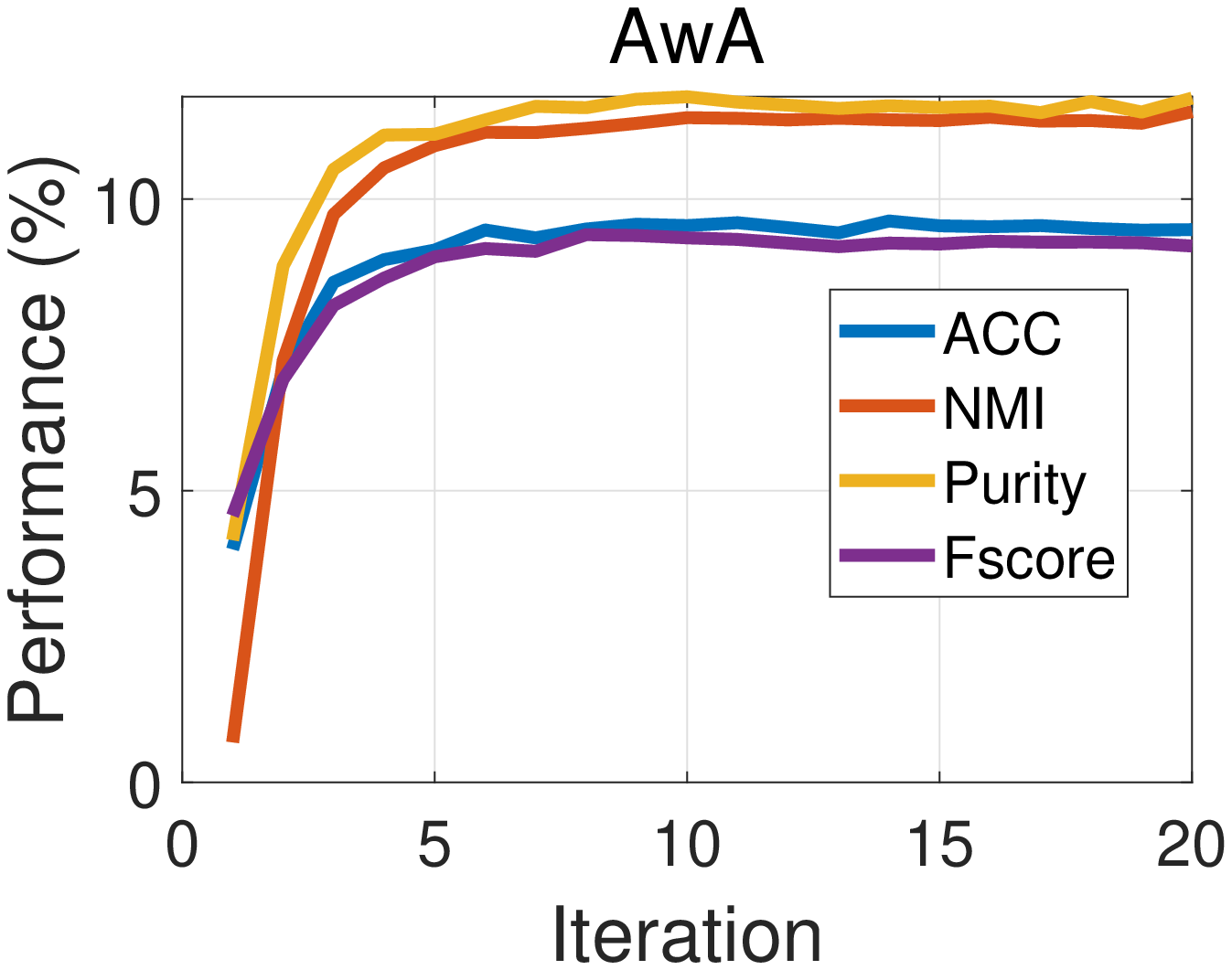}}
\hspace{-0.2cm}
	\subfigure{
		\includegraphics[width=0.23\textwidth]{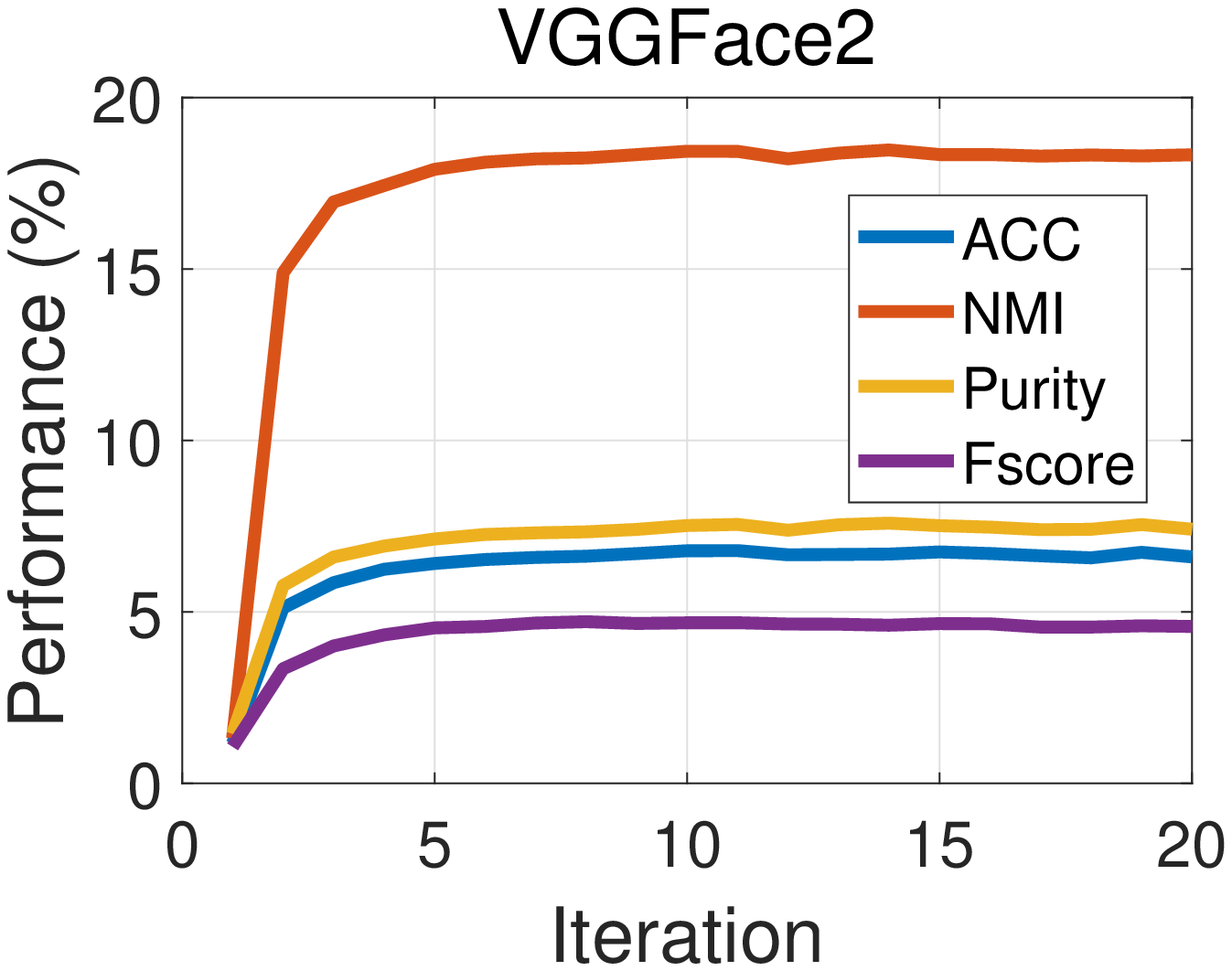}}
\hspace{-0.2cm}	

	\caption{The objective values of AWMVC vary with iterations (above) and the evolution of the consensus coefficient matrix $\mathbf{M}$ (below). The results on other datasets are similar and omitted due to space limitations.}
\label{fig_obj}
\end{figure}
 As proved above, our proposed algorithm is convergent in theoretic. To verify it in practice and discuss the convergence rate, we plot the objective value of AWMVC changes with iterations on AwA and VGGFace2, shown in Figure \ref{fig_obj}. From the figure, it is seen that the objective value decreases monotonously and converges in less than ten iterations. In addition, to investigate the evolution of the consensus coefficient matrix of AWMVC, in each iteration, we take $\mathbf{M}$ as the input to conduct k-means on the same datasets. The results are also shown in Figure \ref{fig_obj}. From the figure, it is obtained that the clustering performance gradually gets better and then stays stable, which shows the effectiveness of our proposed method.

\subsection{Dimension Weights}
\begin{figure}[]
\caption{Dimension weights of AWMVC on seven benchmark datasets. From left to right, the dimensions are $k$,$2k$ and $3k$, respectively.}
	\centering
	
	\subfigure{\includegraphics[width=0.45\textwidth]{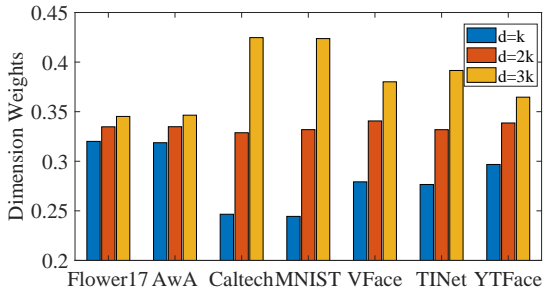}}

	\label{weight_fig}
\end{figure}

We plot the optimized weights for different dimensions on seven datasets in Figure \ref{weight_fig}. Considering that the embedding with higher dimensions contains richer information, their final weights ought to be larger, and the results on all datasets are as expected in our experiments. Consequently, our model can effectually integrate diverse and complementary information among views in a discriminatory way.
\subsection{Running Time Comparison}
\begin{table*}[htbp]
	\centering
	\caption{Running time comparison with state-of-the-art algorithms on seven benchmark datasets.}
	\label{running_time}
	\small 
	\begin{tabular}{|c|c|c|c|c|c|c|c|c|c|c|}
		
    \hline
        Datasets & AMGL & UOMVSC & MNMF & SMVSC & BMVC & LMVSC & FPMVS-CAG & OPMC & AWMVC  \\ \hline
        Flower17 & 378.82  & 192.57  & 176.95  & 77.38  & 1.72  & 2.55  & 100.89  & 45.48  & 42.76   \\ \hline
        AwA & - & - & 3532.20  & 1061.88  & 67.87  & 206.83  & 2834.32  & 693.25  & 434.70   \\ \hline
        Caltech256 & - & - & 1435.04  & 1716.02  & 53.09  & 797.29  & 2588.30  & 212.99  & 796.32   \\ \hline
        MNIST & - & - & 367.46  & 604.59  & 13.85  & 342.26  & 607.07  & 29.07  & 562.31   \\ \hline
        VGGFace2 & - & - & 2526.49  & 2378.45  & 74.53  & 1631.52  & 4058.34  & 499.50  & 1733.50   \\ \hline
        TinyImageNet & - & - & 6464.19  & 7992.69  & 138.87  & 2436.29  & 5870.46  & 972.72  & 3165.95   \\ \hline
        YouTubeFace50 & - & - & 2517.03  & 1818.27  & 98.71  & 1801.96  & 2690.21  & 101.92  & 4160.96   \\ \hline
	\end{tabular}
		
\end{table*}

To compare the computational efficiency of the proposed algorithm, we report the execution time of AWMVC and the compared methods on seven benchmark datasets in Table \ref{running_time}. From the table, it can be seen that:
\begin{enumerate}
\item Compared with graph-based methods such as AMGL and UOMVSC, AWMVC is time-consuming and can handle large-scale situations. Also, AWMVC shows superior clustering performance, demonstrating its efficiency.
\item In comparison to MF-based methods like MNMF and OPMC, AWMVC is parameter-free. MNMF undergoes a hyperparameter and needs to run the implementation multiple times to choose a 'suitable' value. Although OPMC is parameter-free, the initialization significantly impacts the final result. Therefore, the algorithm has to be repeated many times to attain a minor loss. In contrast, we merely perform AWMVC once to get the clustering result. Thus, our proposed method is more time-economic in total.
\item The difference in running time between multi-view subspace clustering and AWMVC is slight. For instance, compared to SMVSC and FPMVS-CAG, AWMVC consumes less time on almost all datasets except YouTubeFace50. Furthermore, AWMVC attains better cluster results. Therefore, the computation cost of AWMVC is more worthwhile.
\end{enumerate}

\subsection{Ablation Study}
Our proposed algorithm factorizes data matrices of multiple views into coefficient matrices under diverse dimensions, then auto-weights them to learn a consensus matrix. To evaluate the validity of our model, we formulate two algorithms for comparison, including AWMVC-P and AWMVC-$\boldsymbol{\alpha}$. To investigate the availability of mapping data matrices into diverse dimensions, we develop the AWMVC-P method, whose embedding number equals $1$ and the corresponding dimension $d=k$. For AWMVC-$\boldsymbol{\alpha}$ method, we allocate each embedding with the same weights to research the influence of the discriminatory factor $\boldsymbol{\alpha}$ on AWMVC. The results of our ablation study are reported in Table \ref{ablation}. The table shows that the clustering performance descends when one of the critical components of AWMVC is dropped out. Therefore, the success of our work is related to the mapping of data to multiple dimensions and the obtained discriminatory information. 

\begin{table}[]
\centering
	\caption{The ablation study of our proposed method on seven benchmark datasets in terms of ACC. The best results are marked in bold.}
	\label{ablation}
	\small 
\begin{tabular}{|c|c|c|c|}
\hline
Datasets      & AWMVC-P      & AWMVC-$\boldsymbol{\alpha}$ & AWMVC          \\ \hline
Flower17      & 46.53         & 48.03    & \textbf{49.12} \\ \hline
AwA           & \textbf{9.44} & 9.32     & 9.42           \\ \hline
Caltech256    & 11.68         & 12.38    & \textbf{12.75} \\ \hline
MNIST         & 96.84         & 98.15    & \textbf{98.85} \\ \hline
VGGFace2      & 6.53          & 6.39     & \textbf{6.84}  \\ \hline
TinyImageNet  & 5.02          & 5.05     & \textbf{5.17}  \\ \hline
YouTubeFace50 & 73.71         & 72.63    & \textbf{75.54} \\ \hline
\end{tabular}
\end{table}

\section{Conclusion}
This paper proposes a novel matrix factorization-based method termed auto-weighted multi-view clustering for large-scale data (AWMVC). Distinct from existing large-scale multi-view clustering algorithms, AWMVC maps data matrices into diverse latent spaces and combines the coefficient matrices from different spaces into a consensus one utilizing discriminatory information. To solve the resultant problem, we develop a six-step alternating algorithm with proven convergence, both theoretical and experimental. The linear complexity and parameter-free property enable it to handle large-scale datasets. Comprehensive experiments demonstrate its superior performance, and the ablation study verifies the effectiveness of the critical components of AWMVC. Besides, mapping data matrices into diverse dimensions can be easily used with existing multi-view clustering methods. In future work, we intend to extend it to handle incomplete views. 

\section{Acknowledgments}
This work was supported by the National Key R$\&$D Program of China 2020AAA0107100 and the National Natural Science Foundation of China (project no. 61872371, 61922088 and 61976196).

\bibliography{aaai23}

\end{document}